\documentclass{article}
\usepackage[utf8]{inputenc} 
\usepackage[T1]{fontenc}    
\usepackage{arxiv}

\usepackage{cite}
\usepackage{amsmath,amssymb,amsfonts}
\usepackage{algorithmic}
\usepackage{graphicx}
\usepackage{textcomp}
\usepackage{breqn}
\DeclareMathOperator*{\argmin}{argmin}

\begin{document}
\title{Stochastic Configuration
Machines: FPGA Implementation}
\author{Matthew J. Felicetti\\
Department of Engineering, La Trobe University\\
Bendigo VIC 3552, Australia
\And
Dianhui Wang
\thanks{Corresponding Author (dh.wang@deepscn.com)}\\
Artificial Intelligence Research Institute, China University of Mining and Technology\\
Xuzhou 221116, China \\
State Key Laboratory of Synthetical Automation for Process Industries\\
Northeastern University, Shenyang 110819, China\\}

\maketitle

\begin{abstract}
Neural networks for industrial applications generally have additional constraints such as response speed, memory size and power usage. Randomized learners can address some of these issues. However, hardware solutions can provide better resource reduction whilst maintaining the model's performance. Stochastic configuration networks (SCNs) are a prime choice in industrial applications due to their merits and feasibility for data modelling. Stochastic Configuration Machines (SCMs) extend this to focus on reducing the memory constraints by limiting the randomized weights to a binary value ${\{-1,1\}}$ with a scalar for each node and using a mechanism model to improve the learning performance and result interpretability. This paper aims to implement SCM models on a field programmable gate array (FPGA) and introduce binary coded inputs to the algorithm. Results are reported for two benchmark and two industrial datasets, including SCM with single-layer and deep architectures. 
\end{abstract}

\section{Introduction}
\label{sec:introduction}

An approach to elevate the computational burden of neural networks and for applications in hardware, is to reduce the amount and accuracy of the physical data. Typically, neural network weights are stored as floating point values, either 32 or 64-bit. Gupta and Narayanan\cite{DBLP:journals/corr/GuptaAGN15} demonstrated that computational improvements could be achieved with good generalization using limited numerical precision, namely 16-bit fixed point values. Further, activation functions require division or many multiplications, which can take up to 10-20 processor cycles per operation. Courbariaux, David and Bengio\cite{courbariaux2014training} show that low-precision multiplications can be sufficient for training neural networks. Courbariaux, David and Bengio \cite{DBLP:journals/corr/CourbariauxBD15} can then demonstrate the use of binary weights during forward and back-propagation, known as BinaryConnect, which in turn can be used for memory reduction and speed enhancements. BinaryConnect presents two binarization techniques of the weights $w$, a deterministic approach based on the sign function and a stochastic approach. The BinaryConnect model is trained using Stochastic Gradient Descent (SGD) with two sets of weights, real and binarized. The real weights are binarized into binary weights to be passed through the forward and backward passes, and the real weights are updated after each pass. Due to the real weights being able to extend past the range of binary values, the real weights are bound between -1 and 1 using a clipping function. BinnaryConnect is further extended in BinaryNet \cite{DBLP:journals/corr/CourbariauxB16} BinaryNet whereby both activation and weights are binarized. A straight-through estimator calculates the gradient of the sign function, allowing it to be used as an activation function. BinaryNet allows for a deep structure where, other than the first layer, all layer inputs are binary. XNOR net\cite{DBLP:journals/corr/RastegariORF16} demonstrates that binary inputs and weights can be used for convolutional layers in convolutional neural networks.
 
 In recent years, randomized neural networks \cite{471375, Pao, doi:10.1002/widm.1200} have received considerable attention due to their ability to speed up training by assigning weights and biases randomly. Stochastic Configuration Networks (SCNs) \cite{8013920} are a class of randomized learners that have demonstrated their ability to outperform other randomized learners. SCNs employ a supervisory mechanism to select random weights and biases and ensure the universal approximation property. The supervisory mechanism generates the weights and biases and scales them dynamically. Some examples of SCNs' promising performance can be seen in \cite{LU2019119}, \cite{wang2017stochastic} and \cite{wang2017robust}. DeepSCN, a multi-layer SCN was introduced in \cite{8489695}, demonstrating that SCN can be applied using a deep structure and can improve the performance over a shallow network. Recent extension to the SCN framework includes an improved selection of hyper-parameters using Monte-Carlo tree search and random search \cite{FELICETTI2022431}, online implementation \cite{li2023online}, and investigation into different distributions of weights \cite{FELICETTI2022819}. To address industrial uses, Stochastic Configuration Machines (SCMs) were introduced in \cite{wang2023stochastic}, where weights are constrained to take binary values, a mechanism model is incorporated and contains an early stopping feature. By using binary values, the model can be very small in terms of memory whilst still outperforming other randomized methods due to the supervisory mechanism.\\
 
In pursuit of faster and smaller neural network models, we see that neural networks are being implemented in hardware. Hardware solutions provide many advantages such as the physical size of the components required, the high-speed computation versus a typical personal computer (PC) \cite{ZHANG2020106}, especially with high numbers of parallel operations, the reduction of power consumption in running the models \cite{6402898, 8456540, 7551399}, and increasing fault tolerance \cite{6402898}. Field programmable gate arrays (FPGAs) are the most commonly used platform for implementation due to being re-configurable. \\

Given SCMs' ability to use the supervisory mechanism to produce quality models that are speed and space efficient, and the opportunities and advantages that exist using FPGAs, we propose introducing a method for implementing SCM models on FPGAs. Our main contributions are as follows:
\begin{enumerate}
    \item Implementation of SCM on an FPGA demonstrating accuracy performance near-identical to a model on a PC.
    \item Introduction of encoding schemes used to encode the input into binary values for SCM. Further, we provide some analysis of preferred encoding schemes for different datasets.
    \item Implementation of FPGA SCM that uses no hardware multipliers by using and introducing different approaches to efficiently calculate the dot product in hardware.
    \item Experimental results demonstrating the speed, power usage and performance.
    \item Demonstration that FPGA SCM can outperform FPGA implementation of SCN.
\end{enumerate}

The remainder of the paper is as follows. Section II overviews SCN, SCM, binarized neural networks and existing FPGA implementations. Section III outlines each part of the SCM FPGA implementation. Section IV evaluates the FPGA model by looking at the input encoding, single-layer performance, deep implementation performance, memory reduction, power, speed and SCN versus SCM. Section V concludes the paper.

\section{Technical Support}

\subsection{Revisit of Stochastic Configuration Machines}

Wang and Li\cite{8013920} have shown that Stochastic Configuration Networks (SCNs) are universal approximators. The SCN framework involves incrementally building a network by randomly assigning the hidden weights and biases with a constrained supervisory mechanism. The output weights can then be evaluated using the least squares method. SCN allows the hidden weights and bias to be evaluated $T_{max}$ times for each iteration to find the most appropriate random values. SCN also employs an adaptive scope setting ($\lambda$), which scales the weights and bias ($w$ and $b$). A deep model further extends this in \cite{8489695}, which is built incrementally over multiple layers. Each hidden layer feeds into the next hidden layer whilst also including connections with output weights to the output layer from each hidden layer.\\

A new randomised learner model based on SCN is recently developed by Wang and Felicetti in \cite{wang2023stochastic}, named stochastic configuration machines (SCMs), designed for industrial applications. SCM stresses the importance of effective modelling whilst reducing data, significantly compressing the models while demonstrating strong prediction performance. SCM adds a mechanism model to DeepSCN; hidden weights take binary values, and an early stopping mechanism is built into the learning algorithm to prevent overfitting. The model can be described mathematically as follows:
\begin{dmath}
Y=P(X,p,u)+\displaystyle\sum_{k=1}^{M}\beta_k H_k(X),
\end{dmath}
where $P(X,p,u)$ represents a mechanism model with a set of parameters $p=[p_1,...,p_l]^T (l\le d)$ and $u=[u_1,...,u_m]$. $
H_k(X)=\phi(W_k^T H_{k-1}(X)+\Theta_k)$ where $\phi(\cdot)$ is an activation function. $\Theta_k=[\theta_1^k,...,\theta_{n_k}^k]^T$ denotes a threshold vector of hidden nodes at the \emph{k-th} layer, $H_0(X)=X$, $k=1,2,..., M$, $n_M=m$ and
\begin{equation}
W_{k} = 
 \begin{bmatrix}
  w_{1,1}^k &  \cdots & w_{1,n_k}^k \\
  \vdots  & \vdots  & \vdots  \\
  w_{n_{k-1},1}^k & \cdots & w_{n_{k-1},n_k}^k 
 \end{bmatrix}_{n_{k-1}\times n_k}
\end{equation}
where $w_{ij}^k$ takes binary values $\{-\lambda,+\lambda\}$, $\lambda\in\{\lambda_1,\lambda_2,...,\lambda_p\}$, representing a random synaptic weight between the \emph{i-th} node at the \emph{k-th} layer (the \emph{0-th} layer is the input layer) and the \emph{j-th} node at the \emph{(k+1)-th} layer (the \emph{L-th} layer is the output layer); $\beta_k$ denotes the readout (or output weight) matrix from the \emph{k-th} hidden layer to the output layer:
\begin{equation}
\beta_{k} = 
 \begin{bmatrix}
  \beta_{1,1}^k &  \cdots & \beta_{1,n_k}^k \\
  \vdots  & \vdots  & \vdots  \\
  \beta_{m,1}^k & \cdots & \beta_{m,n_k}^k 
 \end{bmatrix}_{m\times n_k},
 \beta=\begin{bmatrix}
\beta_1^T \\
\beta_2^T  \\
 \vdots \\
\beta_m^T  
 \end{bmatrix}.
\end{equation}\\

Given a training dataset with $d$ features, $m$ outputs, $N$ training examples, that is, $X_t =[x_{t1}, x_{t2},..., x_{tN}], x_{ti}=[x_{ti,1},...,x_{ti,d}]^T \in \mathbb{R}^d$ and outputs $Y_t = [y_{t1}, y_{t2},..., y_{tN}], y_{ti}=[y_{ti,1},...,y_{ti,m}]^T \in \mathbb{R}^m$. If using a linear regression model for $P(X,p,u)$, the weights are given by $p^*=[p_{1}^*,...,p_{m}^*], p_i^*=[p_{i,1}^*,...,p_{i,d}^*]^T$. 
The residual training error vector before adding the \emph{L-th} hidden node on the \emph{k-th} hidden layer is added, donated by  $\mathcal{E}_{L_k-1}^{(k)} = \mathcal{E}_{L_k-1}^{(k)}(X_t) = [\mathcal{E}_{L_k-1,1}^{(k)}(X_t),...,\mathcal{E}_{L_k-1,m}^{(k)}(X_t)]^T$, where $\mathcal{E}_{L_k-1,q}^{(k)}(X_t) = [\mathcal{E}_{L_k-1,q}^{(k)}(x_{t1}),...,\mathcal{E}_{L_k-1,q}^{(k)}(x_{tN})]^T \in  \mathbb{R}^N, q=1,2,...,m$.
Then, after adding the \emph{L-th} hidden node in the \emph{k-th} hidden layer, we can calculate the output of the \emph{k-th} hidden layer:
\begin{equation}\label{eq:hidden_out}
h_{L_k}^{(k)}:=h_{L_k}^{(k)}(X_t)= [\phi_{k,L_k}(x_1^{(k-1)}),...,\phi_{k,L_k}(x_k^{(k-1)})]^T 
\end{equation}
where $\phi_{k,L_k}(x_i^{(k-1)})$ is used to simplify $\phi_{k,L_k}(x_i^{(k-1)}, w_j^{(k-1)} , b_j^{(k-1)} )$ , and $x_i^{(0)}=x_i=[x_{i,1},...,x_{i,d}]^T, x_i^{(k-1)}=\phi(x^{(k-2)}, W^{(k-1)}, B^{(k-1)})$ for $k \ge 2$. \\
Let $H_{L_k}^{(k)}=[h_1^{(k)},h_2^{(k)},...,h_{L_k}^{(k)}]$ represent the hidden layer output martix. A temporary variable $\xi_{L_k,q}^{(k)} (q=1,2,...,m)$ is introduced: 
\begin{equation}\label{eq:theta}
\xi_{L_k,q}^{(k)}=\frac{\langle \mathcal{E}_{L_k-1,q}^{(k)} , h_{L_k}^{(k)} \rangle^2}{\langle h_{L_k}^{(k)}, h_{L_k}^{(k)} \rangle}-(1-r)\langle \mathcal{E}_{L_k-1,q}^{(k)}, \mathcal{E}_{L_k-1,q}^{(k)}\rangle.
\end{equation}
where  $\left< \cdot,\cdot\right>$ denotes the dot product and the X argument is omitted in $e^{(k)}_{L_k-1,q}$ and $h^{(k)}_{L_k}$.
To calculate the output weights, let $\mathcal{H} = [H_{L_1}^{(1)}, H_{L_2}^{(2)},...,H_{L_M}^{(M)}]\in \mathbb{R}^{N\times \sum_{k=1}^ML_k}$ represent the output matrix of all hidden layer, where $L_k, k=1,2,...,M$, represents the number of nodes at the \emph{k-th} layer, respectively. Then the optimal solution $\beta^*$ is computed using the least squares method as follows:
\begin{equation}\label{eq:beta}
\beta^*= \argmin_{\beta}||\mathcal{H}\beta-(Y_t^T-(X_t^Tp^*+I_1u^*))||^2_F=\mathcal{H}^\dagger (Y_t^T-(X_t^Tp^*+I_1u^*))
\end{equation}
where $\mathcal{H}^\dagger$ is the Moore-Penrose generalised inverse \cite{lancaster1985theory} of the matrix $\mathcal{H}$, $I_1=[1,1,...,1]^T\in\mathbb{R}^N$ and $||\cdot||_F$ denotes the Frobenius norm.\\
The addition of nodes is continued until no suitable candidate nodes can be found or until the early stopping criterion is satisfied. Given a validation dataset with K examples, with inputs $X_v =[x_{v1}, x_{v2},..., x_{vK}], x_{ti}=[x_{vi,1},...,x_{vi,d}]^T \in \mathbb{R}^d$ and outputs $Y_v = [y_{v1}, y_{v2},..., y_{vK}], y_{vi}=[y_{vi,1},...,y_{vi,m}]^T \in \mathbb{R}^m$.
The residual error vector after adding the \emph{L-th} hidden node on the \emph{n-th} hidden layer is donated by  $E_{L_k}^{(k)} = E_{L_k}^{(k)}(X_t)$.
A step size $L_{step}$ and a tolerance $\tau$ are used in the early stopping criterion. If the condition $\frac{E_{L_k-L_{step}}^{(k)}-E_{L_k}^{(k)}}{E_{L_k}^{(k)}}\le\tau$ $(L_k>L_{step})$ is met, then the nodes are iteratively removed until $\frac{E_{L_k-1}^{(k)}-E_{L_k}^{(k)}}{E_{L_k}^{(k)}}>\tau$  is satisfied and nodes for that layer will not be added further. \\
The hidden weights in SCM are defined to be binary, and are scaled by the adaptive scope parameter $\lambda$, resulting in a floating point value. To reduce memory as $\lambda$ is the same value for a given node, the $\lambda$ value is stored for each node such that given the \emph{n-th} layer: 
\begin{equation}
\Upsilon_{n}^*=[\lambda^n_1,\lambda^n_2, ..., \lambda_{L_n}^n], 
\end{equation}
and in turn the weights can be stored in binary and are scaled only when fed forward such that:
\begin{equation}
W_{n}^* = \Upsilon_{n}^*\cdot w_{n}^* .
\end{equation}
For more details on SCM fundamentals, one can refer to \cite{wang2023stochastic}.

\subsection{FPGA implementation concepts}
FPGA have allowed for hardware implementation of neural networks to create high-speed, low-cost networks.  Many implementations have been shown to provide high accuracy and speed \cite{8052266, 8533484, 7857151, 5607329, dicecco2017fpga}. Further, several binarized neural networks have been implemented on an FPGA, including FP-BNN, which uses XNOR and shifts to reduce bottlenecks \cite{LIANG20181072}, recursive binary networks using gradient descent \cite{guan2019recursive} and in convolution networks \cite{zhang2019xnorconv}. 
Issues for neural networks on FPGAs have been investigated in \cite{muthuramalingam2008neural}. The main constraints are the choice between parallel and sequential operations, the precision of bits and non-linear functions. 
\subsubsection{Multiply accumulate operation}
\begin{figure}[!t]
\centering
\includegraphics[width=3.5in]{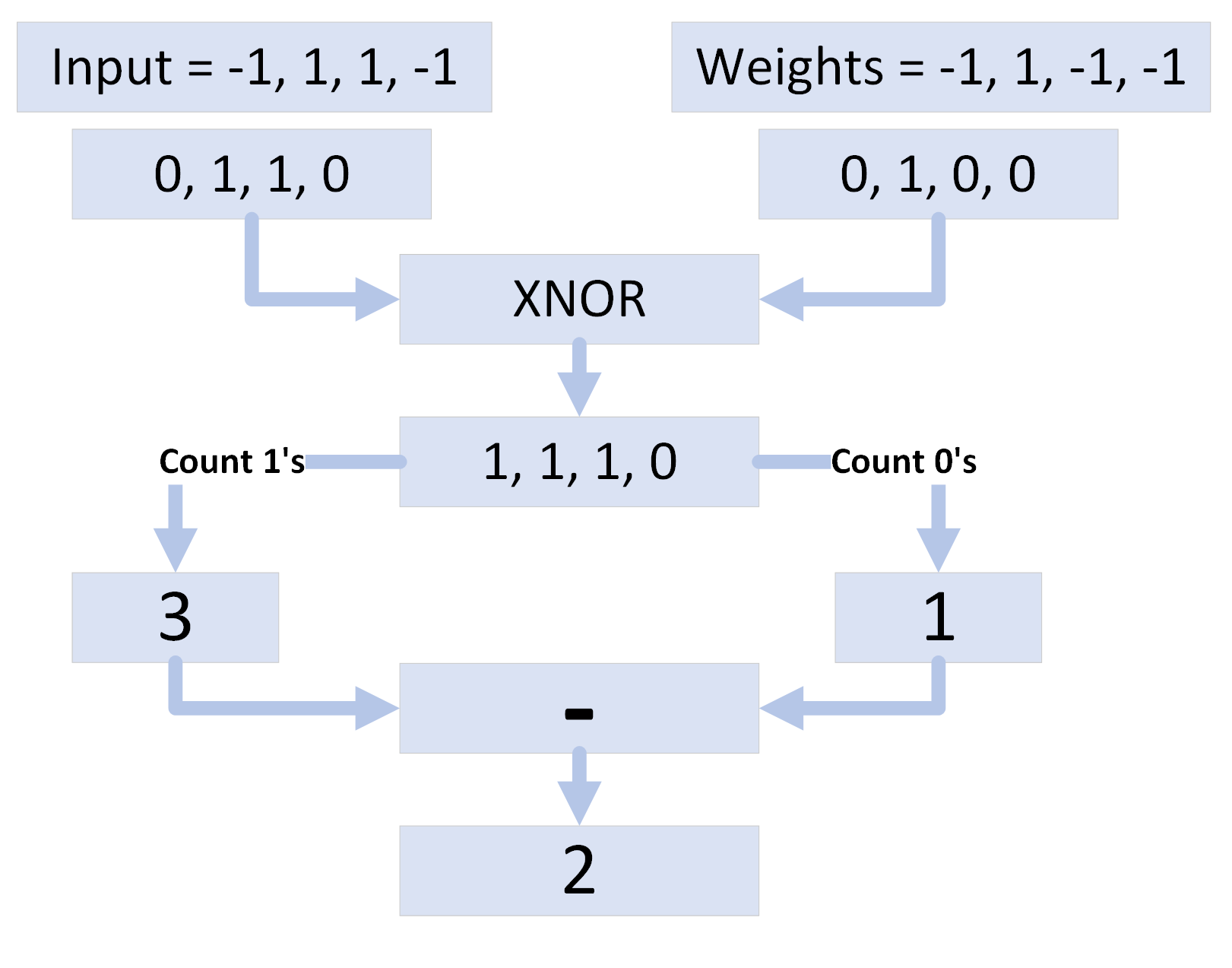}
\caption{XNOR-count operation of two binary vectors with values $\{-1,1\}$. Inputs and weights are firstly mapped from $\{-1,1\}$ to $\{0,1\}$, then an XNOR operation is performed, the '1's and '0' are counted, and the result is the count of '1's minus the count of '0's.  $(-1)\times(-1)+1\times1+1\times(-1)+(-1)\times(-1)=2$ requires four multiplications and four additions, where this method requires only one binary operation, two-bit counts and one subtraction. }
\label{fig:xnor}
\end{figure}

Multiply-accumulate operations can perform the dot product of the weights and layer inputs.  Multiply-accumulate operations can take many multipliers on an FPGA that is resource expensive \cite{1303135}. Several implementations of multiply-accumulate operations have been investigated to approximate the multiply-accumulate\cite{9034956, yang2019approximate, karthikeyan2023energy} or perform more efficient parallel multiply-accumulate operations \cite{8056863, 10.1145/3233300}. Given the case of using binary weights and inputs limited to $\{-1, 1\}$, the multiply accumulate can be completely replaced by a single XNOR count operation  \cite{DBLP:journals/corr/CourbariauxB16}. The XNOR-count operation involves XNORing two binary vectors, then counting the '1's and '0's, and then the resulting value is the count of '0's subtracted from the count of '1's. This requires storing the binary data as '0' and '1', and applying the XNOR-count operation, effectively performing the multiply-accumulate as if the values were '-1' and '1'; this is demonstrated in Figure \ref{fig:xnor}. By doing so, the problem of resources using multiply-accumulate operations is removed. 

\subsubsection{Fixed point encoding}
\begin{figure}[!t]
\centering
\includegraphics[width=3.5in]{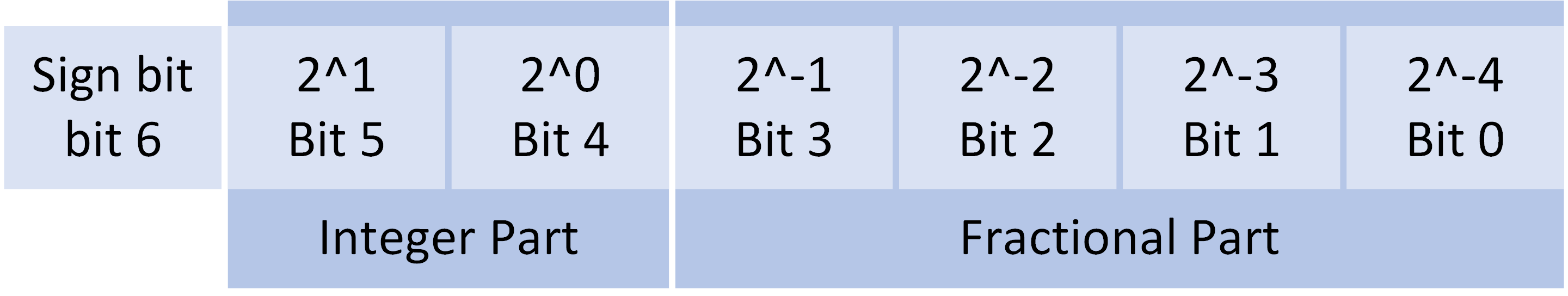}
\caption{Q2.4 fixed point encoding for real values}
\label{fig:fixedpoint}
\end{figure}

Floating point values, such as those defined by the IEEE 754 are widely used to represent real numbers in computing. The two most commonly used floating point types are the IEEE 754 double-precision binary floating-point format, which consists of a sign bit, 11 bits for the exponent and 52 bits for the fraction, and the IEEE 754 single-precision binary floating-point format, which consists of a sign bit, 8 bits for the exponent and 24 bits for the fraction. This standard allows real numbers to be stored and represented accurately. However, for hardware implementation, the floating binary place makes it difficult and requires more hardware resources to implement. Hence, fixed point notation addresses this by storing a fixed number of digits in the fraction. As no exponent exists, a scaling factor needs to be applied to scale the value. Using fixed point values in computation is generally faster and uses fewer resources. The notation for fixed point used in this paper is Q notation, where a signed fixed-point number Qm.f is defined as m integer part bits and f fractional bits; note that the sign bit is not included in this count as shown in Figure \ref{fig:fixedpoint}. Negative numbers are stored in two's complement form. For the application of neural networks where outputs have similar magnitude, fixed-point is a suitable choice\cite{535397}, but care must be taken to prevent overflow, choice of the binary point and the choice of the number of bits to represent the number accurately. The resolution of a fixed point number is half of the precision of the data type. 

\subsubsection{Activation function approximation}
One of the main difficulties of FPGA-based neural networks is representing the activation function. The sigmoid function $\frac{1}{1+e^{-x}}$ is one of the most commonly used activation functions in back-propagation neural networks. However, implementing in hardware is difficult due to the computationally expensive division and exponentiation operations \cite{tommiska2003efficient}.  
Many different look up tables \cite{5642617, ngah2017sigmoid, ortega2014high} and piece-wise approximations \cite{9106311, 8779253} have been proposed. 
\subsubsection{Randomized approaches using FPGA}
More similar to the approach presented in this paper, an implementation of a single hidden layer RVFL network using an FPGA is presented in \cite{9174774}. This implementation uses binary inputs encoded using a density-based encoding scheme to transform a single input into a set of binary inputs. The weights are generated randomly to binary values. The dot product is calculated using the Hadamard product and then summed. The activation used is a clipping function with a configurable threshold. The output of the node is, therefore, an integer. Our implementation is only similar to this in terms of using binary weights and inputs, but the operations between them and the way they are generated is different. \\
In \cite{gao2020fpga} an SCN-based FPGA implementation is proposed. In this implementation, the weight is expressed in terms of its sign and an integer value based on: 
\begin{equation}
    W=sign(W)\times 2^{-r}
\end{equation}
where r is the integer value. Doing so effectively replaces the multiplication into an efficient right-shift operation. This implementation uses the sigmoid activation function, which is approximated using seven different approaches. The three best approaches in this paper are compared with our implementation. The first was the piecewise linear approximation proposed by Tommiska \cite{Tommiska}, the second is the third-order Taylor expansion of the sigmoid function and the third, which performed the best, is introduced in this paper.
A practical application of this and an extended version of an auto-encoder-based SCN is given in \cite{9152009}. Here, the FPGA implementation is used for robotic grasping recognition and demonstrates high recognition accuracy.

\section{FPGA Implementation of SCM}
The following section outlines the implementation details of SCM using an FPGA. The training is performed on a computer using real numbers to produce a model that will be accelerated using an FPGA. All inputs are loaded onto the FPGA, and outputs are sent to a PC using a Universal asynchronous receiver-transmitter (UART) device.
\subsection{Mechanism Model}
The mechanism model for one of the datasets used in this paper is based on machine output; for the remaining three LASSO regression \cite{tibshirani1996regression} is used, \begin{equation} \label{eq:cognitive_chap9}
p_i^*=\argmin_{p_i}\left(\sum^N_{j=1}(y_{tj,i}-\sum^d_{k=1}x_{tj,k}p_{i,k})^2+\alpha \sum^d_{k=1}|p_{i,k}|\right), i=1,2,...,m
\end{equation}
Where $:,i$ denotes all training samples for the i'th output.
And the intercept term of each $p_i^*$ is given by $u^*=\{u_{1}^*,...,u_{m}^*\}$ by finding the mean of each $y_{t:,i}$by finding the mean of each $y_{t:,i}$, that is, $u_i=\frac{\sum^N_{j=1}y_{tj,i}}{N}$, $i=1,2,...,m$.  \\

The inputs are restricted to $\{-1, 1\}$ in the FPGA implementation. With this, the output of the linear model $P(X,p,u)$ can be evaluated by finding the two's complement of the linear weight when the corresponding input is `-1', leaving the value as it is when the corresponding input is `1', then, summing up these values and adding the intercept to get the linear model output. 
\subsection{Input Encoding}
The implementation requires binary inputs to take advantage of XNOR-count operations and binary comparisons used in the selected activation functions. Therefore, the inputs are transformed such that each input becomes a set of binary inputs. Specifically, we require the binary values of -1 and 1 for each input. To perform the XNOR-count operation, the binary input values are stored in FPGA memory as 0 and 1, but they represent -1 and 1, respectively, in calculations. The quantization and representation of the binary inputs will depend on the actual datasets; some will require higher precision and others less. This section outlines density-based encoding and two new encoding schemes for converting inputs into binary inputs.
\subsubsection{Density-based Encoding}
Density-based encoding \cite{9174774} takes a real value in the range [0,1] and a value $N$, where N is the bit vector length. The value is quantized by evenly dividing the value by $N+1$, and then representing the value by an increasing unary code. For example, given an N of 3, the vector can contain "000" for values from $[0,0.25)$, "100" for values from $[0.25,0.5)$, "110" for values from $[0.5,0.75) $ and "111" for value from $[0.75,1]$.
Density-based encoding can provide a binarized representation but does not provide any significance to each bit or set of bits. For high resolution, it would require a very large number of bits. In this application, we use an N of 10. \\
\subsubsection{Input Encoding Scheme 1}
\begin{figure}[!t]
\centering
\includegraphics[width=3.5in]{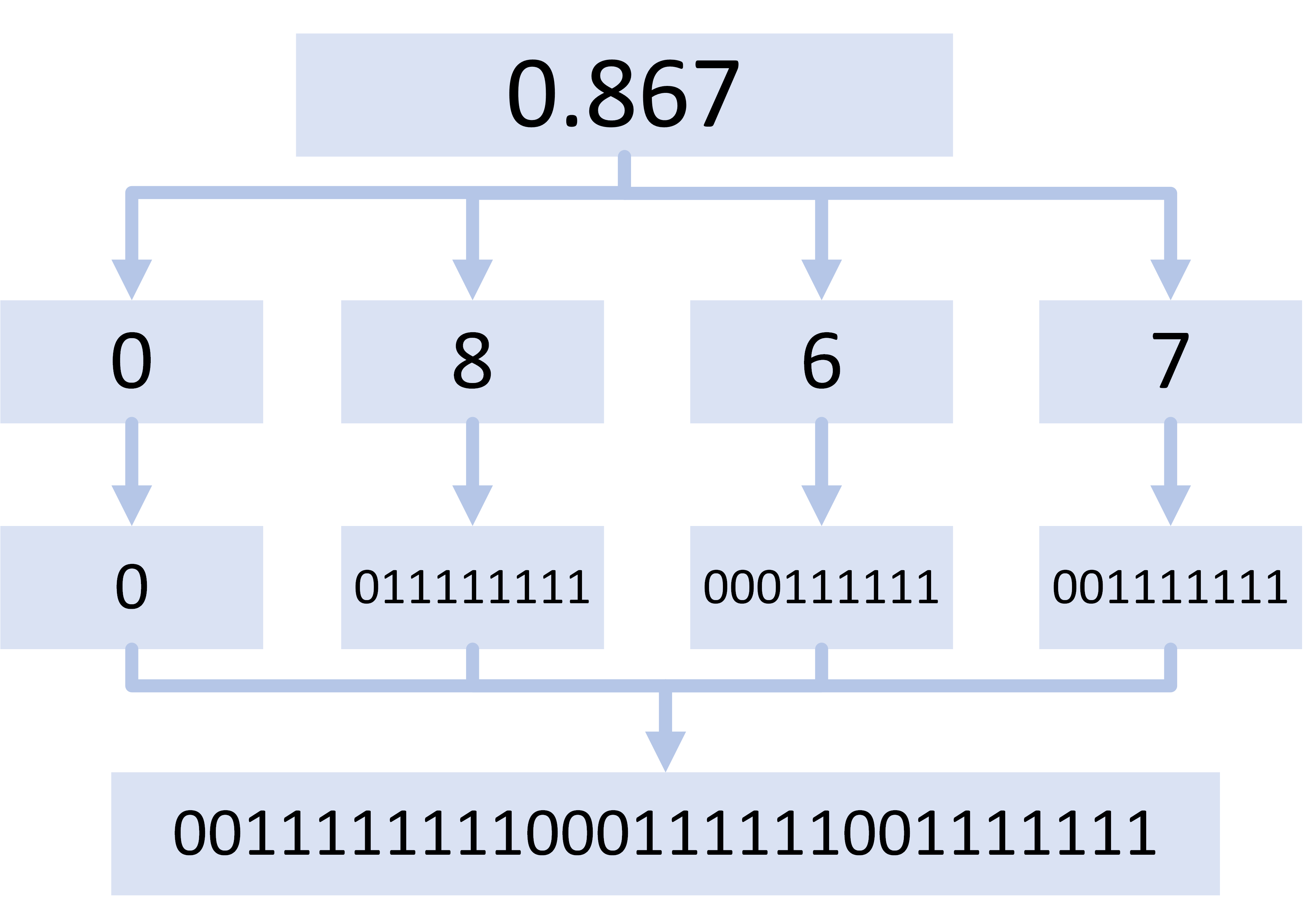}
\caption{Input with value 0.867 using three decimal places being converted into 28 binary inputs using input encoding scheme 1}
\label{fig:encode1}
\end{figure}
Here, we introduce an encoding scheme that breaks the number into its decimal components, and each decimal component is then converted to unary. Inputs are first normalized between 0 and 1. One bit is used to represent the one's place, and then nine bits for each decimal place are converted to its unary representation, left padded with zeros. This can be given as:
\begin{equation}
    x_{i,{bit1}} = x_i\  mod\  0
\end{equation}
\begin{equation}
    x_{i,{bit((k-1)*10)+2\ to\ (k*10))}} = 
    \begin{cases}  000000000 & x\  mod\  10^k = 0 \\ 
    000000001 & x\ mod\  10^k = 1 \\
    000000011 & x\ mod\  10^k = 2 \\
    000000111 & x\ mod\  10^k = 3 \\
    000001111 & x\ mod\  10^k = 4 \\
    000011111 & x\ mod\  10^k = 5 \\
    000111111 & x\ mod\  10^k = 6 \\
    001111111 & x\ mod\  10^k = 7 \\
    011111111 & x\ mod\  10^k = 8 \\
    111111111 & x\ mod\  10^k = 9 \\
    \end{cases}
\end{equation}
where k is the decimal place, b represents the bit number, i is the i'th input, and mod is the modulo function. This scheme allows the number of decimal places to be selected to accommodate finer data. Given $u$ decimal places, this will result in $(1+9*u)*m$ binary inputs, where m is the number of inputs before encoding. Figure \ref{fig:encode1} shows an example of the encoding scheme.
\subsubsection{Input Encoding Scheme 2}
\begin{figure}[!t]
\centering
\includegraphics[width=3.5in]{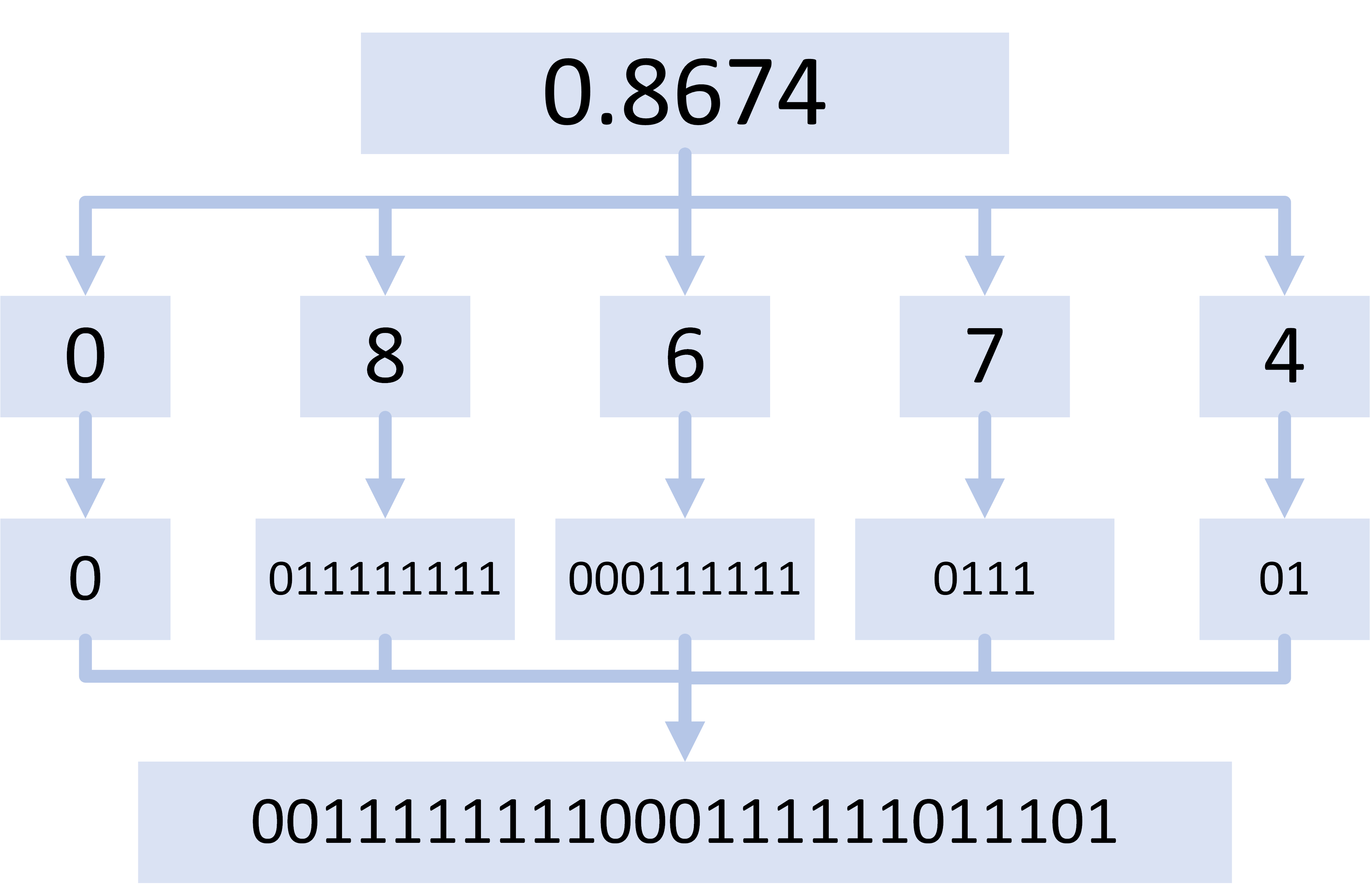}
\caption{Input with value 0.8674 using 4 decimal places being converted into 25 binary inputs using input encoding scheme 2 (V2)}
\label{fig:encode2}
\end{figure}
We know that the leftmost decimal place is the most significant and the rightmost is the least significant; hence we can use this information to reduce the number of binary inputs. Thus, we extend the previous in this encoding scheme by performing further quantization to the lower decimal places. Here, we suggest two alternatives, the first (V1) with one bit for the ones, nine bits for the tenths, four bits for the hundredths and two bits for the thousandths, and the second (V2) one bit for the ones, nine bits for the tenths, nine bits for the hundredths, four bits for the thousandths and two bits for the tens thousandths. 
The first provides a smaller number of binary inputs, with 16 per input, whereas the second provides more accuracy, with 25 binary inputs per input. To convert into four bits the following is used:
\begin{equation}
    x_{i,bits} = 
    \begin{cases}  0000 & x\  mod\  10^k = 0\ or\ 1\\ 
    0001 & x\ mod\  10^k = 2\ or\ 3 \\
    0011 & x\ mod\  10^k = 4\ or\ 5 \\
    0111 & x\ mod\  10^k = 6\ or\ 7 \\
    1111 & x\  mod\  10^k = 8\ or\ 9\\ 
    \end{cases}
\end{equation}
and for two bits, the following is used:
\begin{equation}
    x_{i,bits} = 
     \begin{cases}  00 & x\  mod\  10^k = 0, 1, 2\ or\ 3\\ 
        01 & x\ mod\  10^k = 4, 5\ or\ 6 \\
        11 & x\ mod\  10^k = 7, 8\ or\ 9 \\
        \end{cases}
\end{equation}. Figure \ref{fig:encode2} shows an example of this encoding scheme.
\subsection{Hidden Weights}
The hidden weights are generated as binary values $\{-1,1\}$ during training using the SCM algorithm. On the FPGA, they are stored as binary vectors where each bit represents a weight stored in memory as `0' and `1', where `0' represents `-1' and `1' represents `1'. The scaling factor is stored separately to save memory.
\subsection{Scaling Vector}
The $\lambda$ values used are constrained to integer powers of two, specifically $\{1, 2, 4, 8, 16, 32, 64, 128\}$, intentionally chosen so that the multiplication can be replaced with a fast binary left shift \cite{marchesi1993fast}. Hence on the FPGA, the values stored are the amount of the shift, i.e., $0, 1, 2, 3, 4, 5, 6, 7$, which can be stored efficiently in three bits.
\subsection{Node Implementation}
One advantage of SCN and SCM is the lack of restriction on the activation function, unlike back-propagation, which requires the function to be differentiable. With this, we do not require complex functions like sigmoid that usually require approximations in hardware implementations. Instead, in the application of hardware SCM, we limit the activations to sign and step. \\

In SCM, each node has two outputs, a single bit that will feed into the following layers nodes, and a real value that will feed into the output. SCM calculates these as follows:
\begin{enumerate}
\item Calculates the dot product of the input and hidden weights ($w$).
\item Scales by the scaling factor ($\lambda$).
\item Adds the bias ($b$) to the result.
\item Applies the activation function and output to the following layer.
\item Multiplies the output of the activation function by the output weight ($\beta)$.
\end{enumerate}
\subsubsection{Binarized Dot Product}
\begin{figure}[!t]
\centering
\includegraphics[width=3.5in]{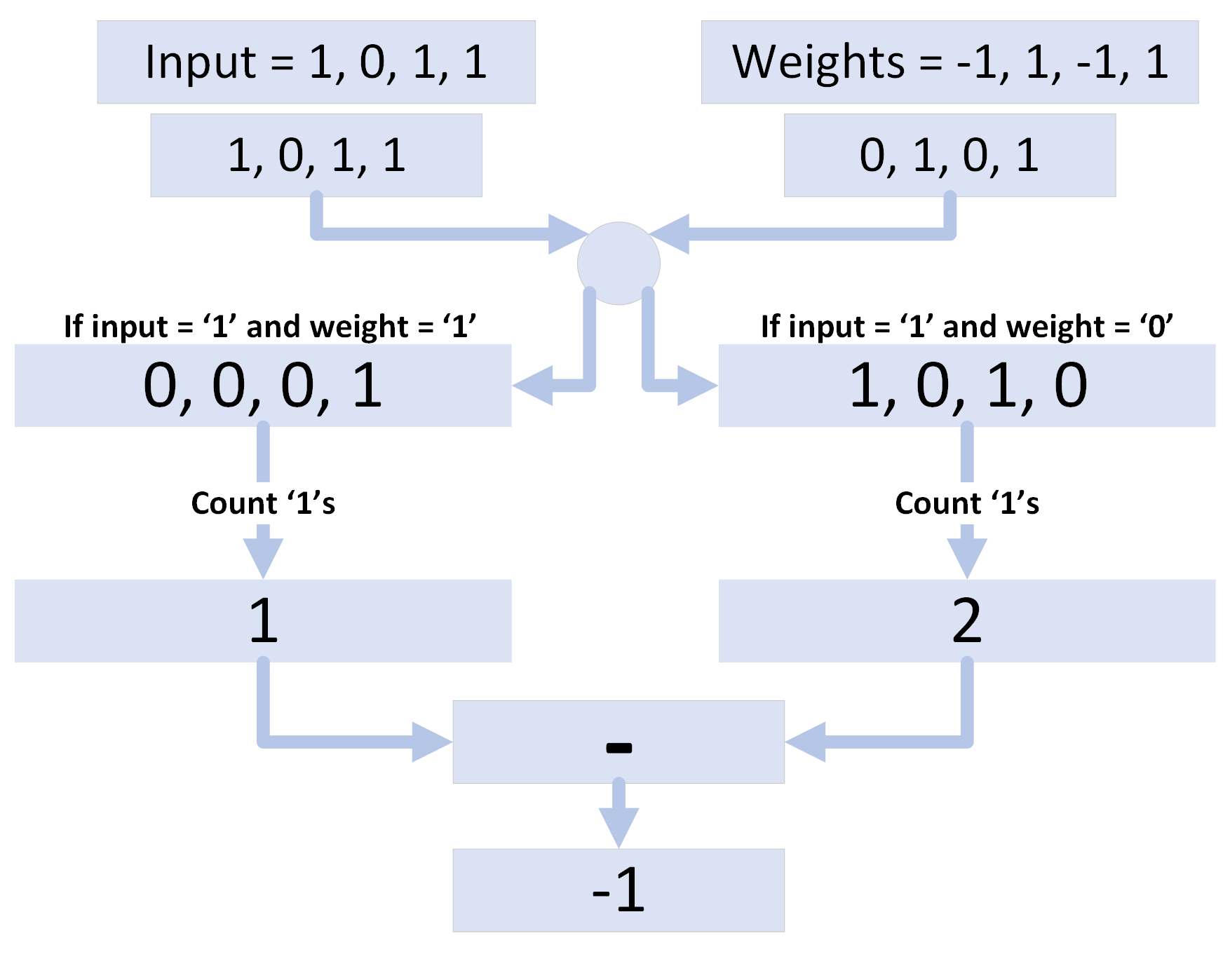}
\caption{{1,0} input and {-1,1} weights dot product operation. Weights are firstly mapped from $\{-1,1\}$ to $\{0,1\}$, then a condition is checked whereby if the input is a `1' and the weight is a `1' a `1' is placed as a flag in the count `1's vector, else if the input is a `1' and the weight is a `0' then a `1' is placed as a flag in the count `-1's vector, the `1's in each vector are counted, and the result is the difference of `1's minus the count of `-1's.  $(1)\times(-1)+0\times1+1\times(-1)+1\times1=-1$ requires four multiplications and four additions, where this method requires two conditions, two-bit counts and one subtraction. }
\label{fig:oneszero}
\end{figure}

 The model's inputs are binary, limited to '-1' and '1', but the output of the activations can be either '-1' or '1' in the case of sign or, '0' or '1' in the case of step. The dot product, therefore, is calculated differently for $\{-1,1\}$ node inputs and $\{0,1\}$ node inputs:
\begin{itemize}
\item With $\{-1,1\}$ inputs (Stored as $\{0,1\}$), the dot product of the weights and inputs can be calculated using the XNOR-count operation detailed in the background.  
\item With $\{0,1\}$ inputs into a node, we only need to consider inputs that are `1' and the respective weights, as inputs that are `0' and their corresponding weight do not contribute to the dot product. For inputs that are `1', the number of weights that are `-1'(stored as `0') and the number of weights that are `1' can be counted, and from this the difference can be calculated (`1's minus `-1's) resulting in the dot product. Practically, this is implemented by having two binary vectors, one for storing `1's and the other for storing `-1's. If the input is `1' and the weight is `1' then a `1' is stored in the inputs position in the `1's vector. If the input is `1' and the weight is `-1' then a `1' is stored in the inputs position in the `-1's vector; otherwise a `0' is stored. This process is summarized in Figure \ref{fig:oneszero}. This operation only requires two conditions per input, two-bit counts and one subtraction.
\end{itemize}
\subsubsection{Sign activation}
The sign activation is implemented using a threshold, whereby if the dot product of the weights and inputs, scaled using lambda and with the bias added, is greater than 0, then the input for the next layer is `1', and the output is $\beta_j^{(k)}$; otherwise, the input to the next layer is `0', and the output is 0. 
\subsubsection{Step activation}
The step activation is similarly implemented using a threshold, and is implemented the same as the sign function for the output bit to the next layer, as the next layer will treat the `0' as a `-1' input. For the real output, if the activation results in a `1', then the output is $\beta_j^{(k)}$, otherwise the two's complement of $\beta_j^{(k)}$ is output, effectively multiplying it by -1. 
\subsection{Output, Linear Weights and Output Weights Representation}
The output, linear weights and output weights are stored in fixed point notation using a 32-bit signed value with Q7.25 encoding and no scaling. With 25 bits associated with the fractional part, the values have a resolution of $\frac{2^{-25}}{2}\approx1.49\times10^{-8}$. 
\subsection{Structure}
\begin{figure}[!t]
\centering
\includegraphics[width=3.5in]{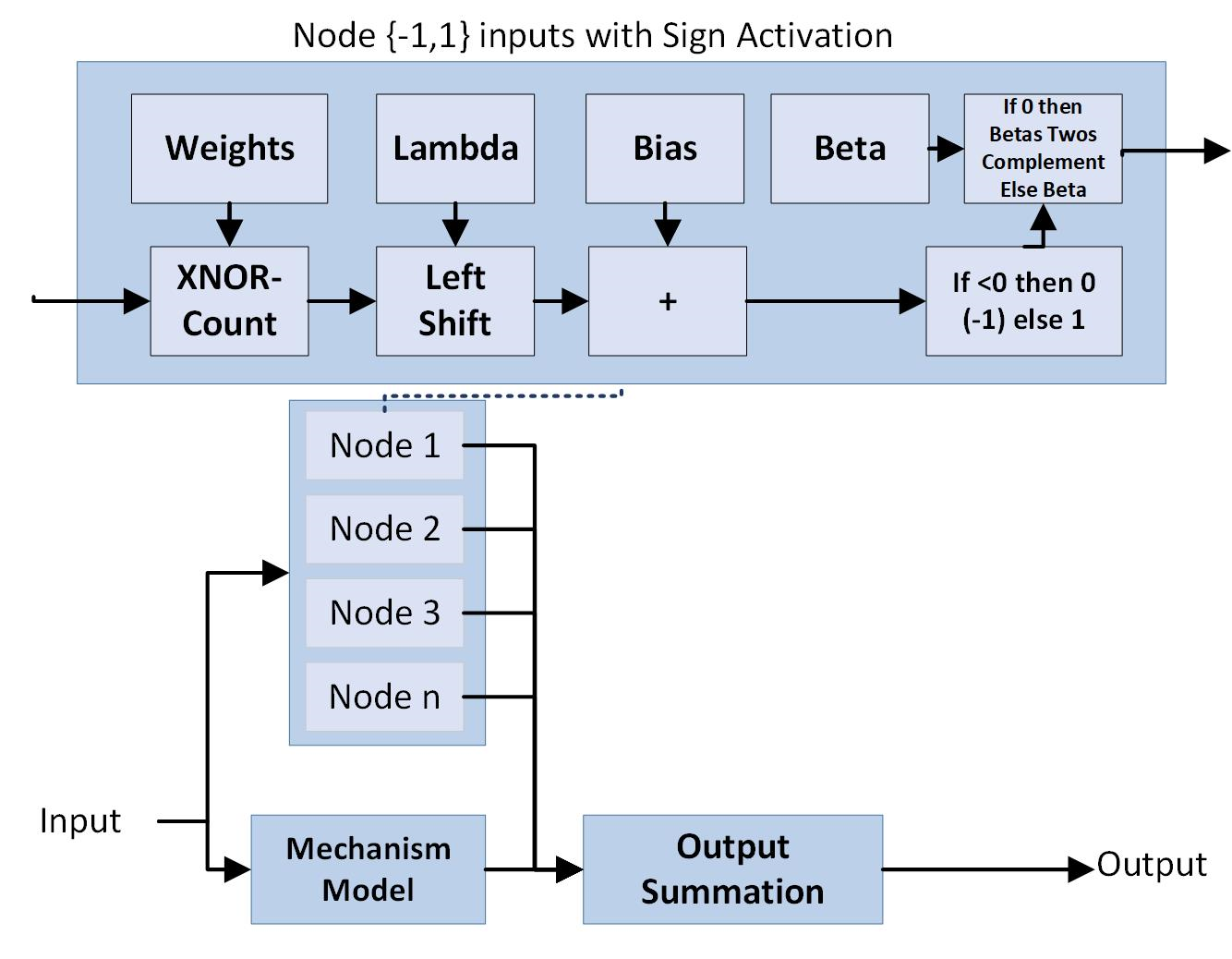}
\caption{Single layer FPGA SCM model structure. The node evaluation with sign activation function is shown for a single node}
\label{fig:single}
\end{figure}
\begin{figure}[!t]
\centering
\includegraphics[width=3.5in]{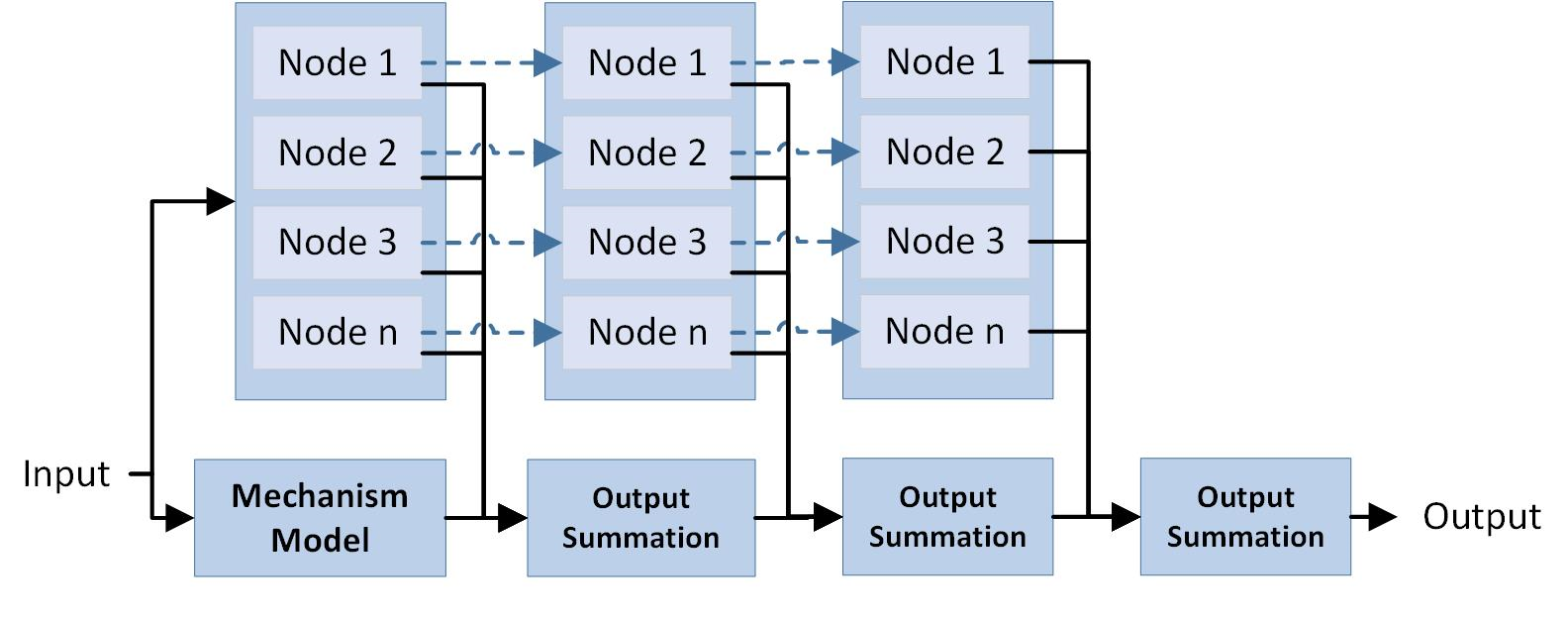}
\caption{Three layer FPGA SCM structure. The output of the activation function feeds into the following node in the next layer, whereas the output feeds into the output summation.}
\label{fig:deep}
\end{figure}
In the single-layer model, the mechanism model and the nodes are all evaluated in parallel. The output of these is then summed to give the model's output. The nodes do not need to output the result of the activation, as there are no further layers to feed into, and all inputs into the nodes are {-1,1}. This is summarized in Figure \ref{fig:single}.The deep model involves the summation of the previous layer in parallel with the node evaluation and is summarized in Figure \ref{fig:deep}. 
\section{Performance Evaluation}
This section gives the performance of different encoding schemes and FPGA models on three datasets. Performance evaluation includes single-layer performance, FPGA SCN versus FPGA SCM, deep implementation performance, memory reduction, power and speed. Training and encoding evaluation was performed in Matlab and run on a PC with an Intel Core i7-3820 @ 3.6GHz and 32GB of RAM. Models were built on an XC7A100T-1CSG324C FPGA with a 100MHz clock. 
\subsection{Datasets}
\begin{itemize}
\item DB1 is a dataset based on the real value function defined as
\begin{equation}
f(x)=0.2e^{-(10x-4)^2}+0.5e^{-(90x-40)^2}+0.3e^{-(80x-20)^2}
\end{equation}
The dataset contains 1300 instances generated from the uniform distribution [0,1], 1000 instances are used in training and 300 in testing.

\item DB2 is a dataset generated from the Rastrigin function \cite{10.1007/BFb0029787} defined as 
\begin{equation}
f(x)=An+\sum_{i=1}^{n}[x_i^2-A\cos{2\pi x_i}]
\end{equation}
where $A=10$, $x_i\in[-5.12,5.12]$. A $n$ of 2 is used, with 40000 training instances and 4489 testing instances, and the inputs are normalized between 0 and 1. 
\item DB3 is a real industrial dataset of the exit thickness in the hot rolling metal process. Hot rolling is a process where rollers apply force on a hot metal strip to produce a metal strip of a desired thickness. The tuning of the industrial mechanisms parameters is usually performed by a mechanism prediction model (MPM) for determining the thickness of a rolled piece and can be expressed as:\\
\begin{equation}
h=S-(\delta_F+\delta_{wear}+\delta_{exp}+\delta{adapt})
\end{equation}
where h is the predicted thickness of the rolling piece, S the roll gap, and the compensation models: $\delta_F$ the rolling mill bounce, $\delta_{wear}$ the roll wear, $\delta_{exp}$ the toll thermal expansion and $\delta_{adapt}$ the self-learning model. The MPM suffers from many assumptions and is only based on measurable factors on the mechanism itself. 
The dataset DB3 was obtained from sensors from steel production plants using 12 different target thicknesses. 3163 samples are used, 2863 in the training and 300 in testing. Feature selection involved using grey relational analysis to find the most suitable inputs. The inputs include eight roll gap measurements, entrance thickness, entrance temperature, exit temperature, strip width, eight rolling force measurements, and eight roll linear speed measurements. Further, eight inputs are calculated based on the mechanism model for roll wear. The target is the measurement of the strip thickness. All inputs are normalized between 0 and 1.

\item The dataset DB4, similar to DB3, concerns hot-rolling of steel; however, it aims to predict the rolling force for plates with varying thicknesses. This dataset includes 14 input parameters, including entrance thickness, exit thickness, entrance width, rolling speed, temperature, various measurements of the content of particular elements, and plan view pattern control parameters. Notably, this dataset includes a mechanism model based on key parameters in the production process; this is used to demonstrate the benefit of a mechanism model used with FPGA SCM.

\end{itemize}
\subsection{Input Encoding Performance}
\begin{table}[htbp]
  \centering
  \caption{Input encoding performance simulations}
   \begin{tabular}{cccccc}
    \hline
    \textbf{DB} & \textbf{Encoding}$^a$ & \textbf{Nodes} & \textbf{Binary Inputs} & \textbf{Training } & \textbf{Testing}\\
    \hline
    \hline
    \textit{1} & \textit{None} & \textit{300} & - & \textit{0.00364} & \textit{0.00421}\\
    \hline
    1 & D, N=10 & 300 & 10 & 0.06737 & 0.06688 \\
    1 & S1, u=3 & 300 & 28 & \textbf{0.01092} & 0.01761 \\
    1 & S1, u=4 & 300 & 37 & 0.02994 & 0.03494 \\
    1 & S2, V1 & 300 & 16 & 0.03137 & 0.03086 \\
    1 & S2, V2 & 300 & 25 & 0.01375 & \textbf{0.0161} \\
    \hline
    \textit{2} & \textit{None} & \textit{300} &  - & \textit{0.14724} & \textit{0.14724} \\
    \hline
    2 & D, N=10 & 300 & 20 & 0.11963 & 0.12011\\
    2 & S1, u=3 & 300 & 56 & \textbf{0.01781} & \textbf{0.02222} \\
    2 & S1, u=4 & 300 & 74 & 0.02281 & 0.02887 \\
    2 & S2, V1 & 300 & 32 & 0.04154 & 0.04352 \\
    2 & S2, V2 & 300 & 50 & 0.019 & 0.0235 \\
    \hline
    \textit{3} & \textit{None} & \textit{60} &  - & \textit{0.00529} & \textit{0.0054} \\
    \hline
    3 & D, N=10 & 60 & 360 & 0.00808 & 0.00898 \\
    3 & S1, u=3 & 60 & 1008 & 0.00715 & 0.00826 \\
    3 & S1, u=4 & 60 & 1332 & 0.00725 & 0.00813 \\
    3 & S2, V1 & 60 & 576 & \textbf{0.0068} & \textbf{0.00785} \\
    3 & S2, V2 & 60 & 900 & 0.00695 & 0.00796 \\
    \hline
    \textit{4} & \textit{None} & \textit{25} &  - & \textit{0.02191} & \textit{0.021528} \\
    \hline
    4 & D, N=10 & 25 & 140 & 0.022283 & 0.023023 \\
    4 & S1, u=3 & 25 & 392 & 0.022302 & 0.022275 \\
    4 & S1, u=4 & 25 & 518 & 0.022476 & 0.023591 \\
    4 & S2, V1 & 25 & 224 & \textbf{0.022028} & \textbf{0.021978} \\
    4 & S2, V2 & 25 & 350 & 0.022424 & 0.022199 \\
    \hline
    \multicolumn{6}{c}{$^a$ D - density based, S1 - Scheme 1, S2 - Scheme 2}
    
    \end{tabular}%

  \label{tab:inputencoding}%
\end{table}%
To understand and choose the appropriate input encoding scheme, each dataset was tested and simulated using real values, density-based encoding and the two encoding schemes introduced in this chapter. Three hundred hidden nodes were used for DB1 and DB2 and sixty for DB3. Five hundred candidates, and the step activation is used for all the datasets. The RMSE error is reported for both training and testing.\\

As shown in Table \ref{tab:inputencoding}, we can observe that the encoding performance is dependent on the dataset. As expected, there is some loss in performance from the encoding. For DB1 for training, scheme 1 with three decimal places performs best, but for testing, scheme 2 (V2) performs best. Density-based encoding performs significantly worse using DB1 and DB2. For DB2, scheme 1 with three decimal places performs best, and scheme 2 (V1) performs worse, demonstrating that there is not enough information in the encoding for this dataset. For DB3, scheme 2 (V1) performs best, which interestingly has the least number of binary inputs other than density-based encoding. For DB4, scheme 2 (V1) performs best for both training and testing. Given these performances, for the FPGA implementation, scheme 2 (V2) is used for DB1, scheme 1 with three decimal places is used for DB2 and scheme 2 (V1) is used for DB3 and DB4. 

\subsection{Single layer implementation performance}
\begin{table}[htbp]
  \centering
  \caption{FPGA and PC based SCM single layer models}
    \begin{tabular}{p{7pt}p{50pt}p{50pt}p{50pt}p{20pt}p{30pt}}
    \hline
    \textbf{DB} & \textbf{FPGA SCM} & \textbf{ PC SCM} & \textbf{Difference} & \textbf{Nodes} & \textbf{Activation} \\
    \hline
    \hline
    1 & 0.040152321 & 0.040147975 & 4.345E-06 & 60 & step \\
    1 & 0.037941259 & 0.037937773 & 3.485E-06 & 60 & sign \\
    2 & 0.03433228 & 0.034332281 & 8.611E-10 & 60 & step \\
    2 & 0.034577532 & 0.034577522 & 1.073E-08 & 60 & sign \\
    3 & 0.008188957 & 0.008206443 & 1.748E-05 & 20 & step \\
    3 & 0.008155897 & 0.008145349 & 1.054E-05 & 20 & sign \\
    4 & 0.021978425 & 0.021978410 & 1.488E-08 & 25 & step \\
    4 & 0.022211285 & 0.022211299 & 1.384E-08 & 25 & sign \\
    \hline
    \end{tabular}%
  \label{tab:singlelayer}%
\end{table}%
For the single-layer performance, six models were built. Each activation function is tested for each dataset. \\

Table \ref{tab:singlelayer} shows the RMSE error for the FPGA and PC-based models and the difference between them.  The sign activation performs better for DB1 and DB3, whereas the step activation performs better for DB2 and DB4. This demonstrates that the activation function chosen is dataset-dependent. Further, it is clear that regardless of activation, the difference between the FPGA and PC models is very small, demonstrating that the quantization selected is suitable for persevering accuracy. \\

\begin{figure*}[!t]
\centering
\includegraphics[width=6in]{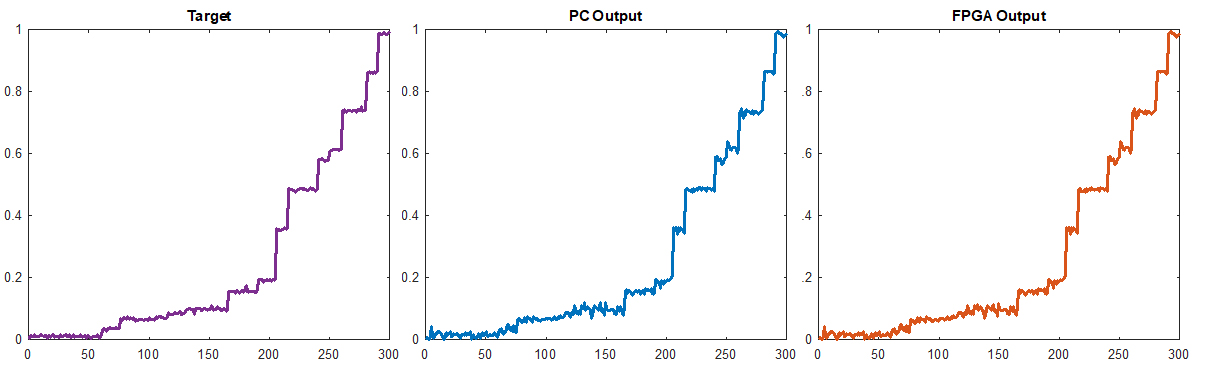}
\caption{Output of target, PC and FPGA of DB3 using single layer SCM with 60 nodes }
\label{fig:I5_fig}
\end{figure*}

Figure \ref{fig:I5_fig} shows the modelling accuracy on DB3, plotting the target, the FPGA output and the PC output. As can be seen, the FPGA and PC outputs are near-identical, demonstrating that the FPGA can produce a model of performance similar to a PC. Further, it can be seen that the model does closely match the target.

\subsection{Deep Implementation Performance}
Different combinations of layers and activations are explored for deep implementation, given in \ref{tab:deep9}. Table \ref{tab:deep9} shows the RMSE error for the FPGA and PC models, which as seen with the single layer implementations, the performance difference between FPGA and PC are minimal. For some combinations, the performance is worse than the single layer; this shows that parameter selection is still critical and dependent on the dataset.\\

\begin{figure}[]
\centering
\includegraphics[width=3.4in]{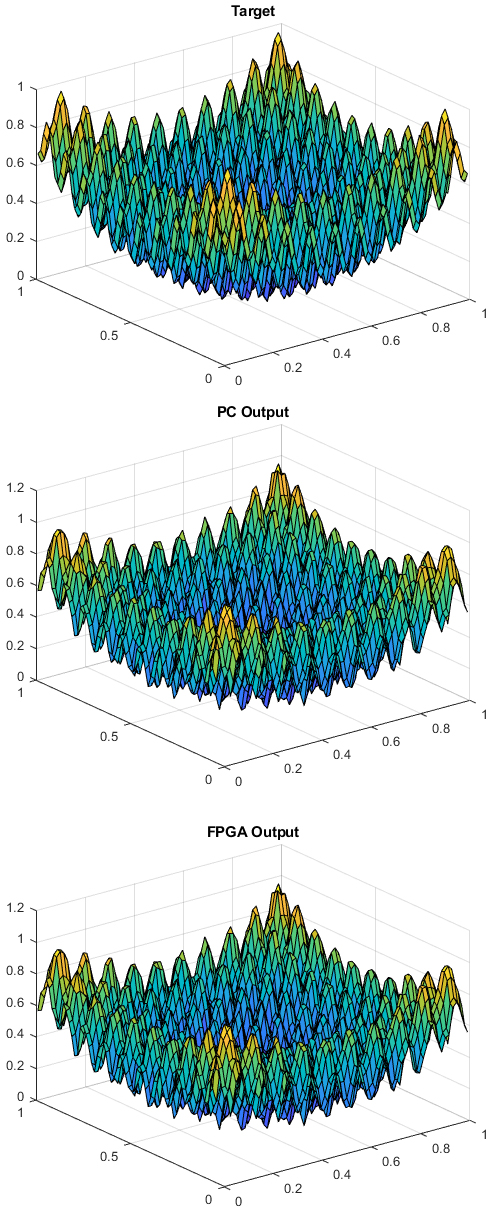}
\caption{Output of target, PC and FPGA of DB2 using deep SCM with 40-40-40 nodes and sign-sign-sign activation functions}
\label{fig:R6_fig}
\end{figure}

Figure \ref{fig:R6_fig} depicts the target, PC output and FPGA output of DB3 using the three-layer model with sign activation. As can be seen, the FPGA can model a complex function with high accuracy and performs similarly to the PC model. 

\begin{table*}[htbp]
  \centering
  \caption{RMSE comparison between FPGA SCM and PC SCM deep models}
    \begin{tabular}{cccccc}
    \hline
    \textbf{DB} & \textbf{FPGA_SCM} & \textbf{PC SCM} & \textbf{Difference} & \textbf{Nodes} & \textbf{Activation} \\
    \hline
    \hline
    1 & 0.031002594 & 0.031001048 & 1.546E-6 & 40-40-40 & sign-sign-sign \\
    1 & 0.032563862 & 0.032563891 & 2.8993E-8 & 60-60 & step-step \\
    2 & 0.033838503 & 0.033838489 & 1.4064E-8 & 40-40-40 & sign-sign-sign \\
    2 & 0.034016416 & 0.034008355 & 8.0609E-6 & 40-40-40 & step-step-step \\
    3 & 0.008418021 & 0.00836614 & 5.1881E-5 & 20-20-20 & sign-sign-sign \\
    3 & 0.008324657 & 0.008324672 & 1.4882E-8 & 20-20 & sign-step \\
     4 & 0.021841345 & 0.021841358 & 1.3789E-8 & 20-8 & sign-sign \\
     4 & 0.022068163 & 0.022073924 & 5.76099E-6 & 20-18 & step-step \\ 
    \hline
    \end{tabular}%
  \label{tab:deep9}%
\end{table*}%

\subsection{Memory Reduction}

\begin{table}[htbp]
  \centering
  \caption{Input memory reduction using FPGA SCM}
    \begin{tabular}{p{10pt}p{30pt}p{40pt}p{30pt}p{30pt}p{50pt}}
    \hline
    \textbf{DB} & \textbf{Inputs (Real)} & \textbf{Inputs (FPGA SCM)} & \textbf{Bits (Real)} & \textbf{Bits (FPGA SCM)} & \textbf{Memory Reduction}\\
    \hline
    \hline
    1 & 1 & 25 & 64 & 25 & 60.9\%\\
    2 & 2 & 56 & 128 & 56 & 56.3\% \\
    3 & 36 & 576 & 2304 & 576 & 75\% \\
    4 & 14 & 224 & 896 & 224 & 75\% \\
    \hline
    \end{tabular}%
  \label{tab:meminputs}%
\end{table}%

\begin{table*}[htbp]
  \centering
  \caption{Weight memory reduction using FPGA SCM}
    \begin{tabular}{crccccc}
    \hline
    \textbf{DB} & \multicolumn{1}{l}{\textbf{Nodes}} & \multicolumn{1}{p{8.57em}}{\textbf{Weights\newline{}(Real)}} & \multicolumn{1}{p{5.355em}}{\textbf{Weights\newline{}(FPGA SCM)}} & \multicolumn{1}{p{6.215em}}{\textbf{Bits\newline{}(Real)}} & \multicolumn{1}{p{10.43em}}{\textbf{Bits\newline{}(FPGA SCM)}} & \multicolumn{1}{p{5.785em}}{\textbf{Memory\newline{} Reduction}} \\
    \hline
    \hline
    1 & 60 & 60 & 1500 & 3840 & 1500 & 60.9375\% \\
    2 & 60 & 120 & 3360 & 7680 & 3360 & 56.25\% \\
    3 & 20 & 720 & 11520 & 46080 & 11520 & 75\% \\
    4 & 25 & 350 & 5600 & 22400 & 5600 & 75\% \\
    \hline
    \end{tabular}%
  \label{tab:weights}%
\end{table*}%

One of SCM's notable features is the reduced memory required to store the weights. Further, on the FPGA, the memory of the inputs can be reduced using an input encoding scheme. And the output weights stored in fixed point offer another area of memory saving. In this analysis, we assume that these would be 64-bit floating-point values, if real values were used. \\

We can see in Table \ref{tab:meminputs} the number of inputs used if a real value was used vs. the number in SCM. It is clear that by encoding the input into bit inputs, the number of inputs increases, but in terms of memory stored, the number of bits used for the input is reduced. For DB3 and DB4, we see a 75\% reduction in input storage memory, compared to using real values due to the encoding scheme. Similarly, we see significant decreases in memory in the other two datasets, thus demonstrating that the encoding schemes do indeed reduce the memory required.\\ 

In Table \ref{tab:weights}, we see the same reduction in memory for the weights in a single-layer model. This is because the number of weights is determined by the number of nodes and the number of inputs. As the number of nodes is the same in this analysis, the number of inputs changes the number of weights, causing the same reduction. This demonstrates that using the encoding schemes and binarized weights results in a significant reduction of memory storage for the weights, and the more significant the reduction in input encoding, the greater the memory reduction.\\

The output weights are encoded using 32-bit signed Q7.25 fixed-point notation; compared to 64-bit floating point, this is a 50\% storage memory reduction for every model. 
\subsection{Power}
The power consumption of a given model was estimated using the Vivado power analysis and optimisation tool. The power estimation gives an idea of the model's power consumption whilst running and includes the communication of outputs via UART. \\

Table \ref{tab:power} shows twelve different models using all three datasets, the number of nodes, the activation, the number of inputs, binary weights, output weights and the power consumption. The single-layer models for each dataset draw around the same amount of power for both sign and step activation. With a larger number of inputs, nodes and weights, the power increases as expected. With the deep models, it is clear that using multiple sign activations vs using multiple step activations is more power-efficient. The reason for this is that the hardware for the XNOR-count operation required for $\{-1,1\}$ inputs into the deeper layers requires less hardware than the $\{1,0\}$ inputs due to the many comparisons.  
\begin{table}[htbp]
  \centering
  \caption{FPGA power consumption}
    \begin{tabular}{p{7pt}p{40pt}p{58pt}p{20pt}p{25pt}p{25pt}p{25pt}}
    \hline
    \textbf{DB} & \textbf{Nodes} & \textbf{Activation} & \textbf{Inputs} & \textbf{Hidden} \par \textbf{Binary} \par \textbf{Weights} & \textbf{Output} \par \textbf{Weights} & \textbf{Power} \\
    \hline
    \hline
    1 & 60 & step & 25 & 1500 & 60 & 0.133W \\
    1 & 60 & sign & 25 & 1500 & 60 & 0.144W \\
    1 & 40-40-40 & sign-sign-sign & 25 & 4200 & 120 & 0.229W \\
    1 & 60-60 & step-step & 25 & 5100 & 120 & 0.4W \\
    2 & 60 & step & 56 & 3360 & 60 & 0.175W \\
    2 & 60 & sign & 56 & 3360 & 60 & 0.187W \\
    2 & 40-40-40 & sign-sign-sign & 56 & 5440 & 120 & 0.262W \\
    2 & 40-40-40 & step-step-step & 56 & 5440 & 120 & 0.463W \\
    3 & 20 & step & 576 & 11520 & 20 & 0.394W \\
    3 & 20 & sign & 576 & 11520 & 20 & 0.391W \\
    3 & 20-20-20 & sign-sign-sign & 576 & 12320 & 60 & 0.428W \\
    3 & 20-20 & sign-step & 576 & 11920 & 40 & 0.459W \\
    4 & 25 & step & 240 & 5600 & 25 & 0.262W \\
    4 & 25 & sign & 240 & 5600 & 25 & 0.262W \\
    4 & 20-8 & sign-sign & 240 & 4640 & 28 & 0.241W \\
    4 & 20-18 & step-step & 240 & 4840 & 28 & 0.263 \\
    \hline
    \end{tabular}%
  \label{tab:power}%
\end{table}%
\subsection{Speed}
The main advantage of using an FPGA with SCM is increased speed performance. In this implementation, there are no time-consuming multiplications; instead, these are replaced with bit shifts, XNOR, count, comparisons, and two's complement operations. Activations are replaced with highly efficient activation functions based on a threshold to determine the output. Additions are carried out on fixed-point values with no scaling, resulting in quick summation. \\

The time for the model to evaluate one input is shown in Table \ref{tab:time9}. This table shows that for single-layer networks, it takes nine clock cycles for DB1 and ten clock cycles for DB2, 3 and 4. The reason for the additional clock cycle is that due to DB2 and DB3 having more inputs, the addition is done over two clock cycles using addition hierarchy. This follows with the two-layer network for DB1 taking 18 clock cycles and 19 clock cycles for DB3 and DB4. For a three-layer network for DB1 it takes 23 clock cycles and 24 for DB2 and DB3. The number of nodes does not affect the speed of these datasets, but if a large number of nodes were added, only a small amount of clock cycles would need to be added to allow for the other additions. With a clock speed of 100MHz, each evaluation of an input is less than 250ns, which could be dramatically sped up with a faster FPGA clock if needed.\\

To further evaluate the speed, the steps at each clock cycle for DB1 using a single layer with the sign activation are summarized below:
\begin{enumerate}
    \item Input is loaded.
    \item Mechanism weights if the corresponding input is `0', then the value is inverted. XNOR is calculated for the inputs and weights for each node.
    \item Mechanism part is summed, and the intercept added. XNOR results are counted into sum of `1's and `0's for each node.
    \item Difference between the sum of `1's and `0's counted for each node.
    \item The difference is left shifted by the $\lambda$ value for each node.
    \item The bias is added to the shifted value for each node.
    \item If the value is less than 0, then output 0; else if 1, then output $\beta$ for each node.
    \item Output of the nodes are added such that the first twenty are summed together, the second lot of twenty are summed together, and the third lot of twenty are summed together.
    \item The three sums and the mechanism part are added; this is the output.
\end{enumerate}
As can be seen in the steps in the clock cycles above, the addition of the output is done over two clock cycles to meet timing requirements; this is done wherever there are many additions. 

\begin{table}[htbp]
  \centering
  \caption{Clock cycles required to evaluate a single input using FPGA SCM}
    \begin{tabular}{ccccc}
    \hline
    \textbf{DB} & \textbf{Clock cycles} & \textbf{Nodes} & \textbf{Activation} & \textbf{Time @100MHz} \\
    \hline
    \hline
    1 & 9 & 60 & step & 90ns \\
    1 & 9 & 60 & sign & 90ns \\
    1 & 23 & 40-40-40 & sign-sign-sign & 230ns \\
    1 & 18 & 60-60 & step-step & 180ns \\
    2 & 10 & 60 & step & 100ns \\
    2 & 10 & 60 & sign & 100ns \\
    2 & 24 & 40-40-40 & sign-sign-sign & 240ns \\
    2 & 24 & 40-40-40 & step-step-step & 240ns \\
    3 & 10 & 20 & step & 100ns \\
    3 & 10 & 20 & sign & 100ns \\
    3 & 24 & 20-20-20 & sign-sign-sign & 240ns \\
    3 & 19 & 20-20 & sign-step & 190ns\\
    4 & 10 & 25 & step & 100ns \\
    4 & 10 & 25 & sign & 100ns \\
    4 & 19 & 20-8 & sign-sign & 190ns \\
    4 & 19 & 20-18 & step-step & 190ns \\
    \hline
    \end{tabular}%
  \label{tab:time9}%
\end{table}%

\subsection{SCN vs SCM}
\begin{table}[htbp]
  \centering
  \caption{FPGA SCN versus FPGA SCM}
    \begin{tabular}{p{10pt}p{37pt}p{36pt}p{28pt}p{26pt}p{24pt}p{24pt}p{44pt}}
    \hline
    \textbf{DB} & \textbf{Algorithm} \par \textbf{(FPGA)} & \textbf{Activation} & \textbf{RMSE} & \textbf{Power (Watts)} & \textbf{Clock} \par \textbf{Cycles} & \textbf{Nodes} & \textbf{Time} \par \textbf{@100MHz} \\
    \hline
    \hline
    1 & SCN\cite{gao2020fpga} & Sigmoid & 0.0551 & 0.991$^\dagger$ & 18 & 18 & 180ns* \\
    1 & SCM & Step & \textbf{0.0531} & 0.112$^\dagger$ & 9 & 18  & 90ns \\
    1 & SCM & Sign & 0.0538 & 0.114$^\dagger$ & 9 & 18  & 90ns \\
    1 & SCM & Step & 0.0401 & 0.133$^\dagger$ & 9 & 60  & 90ns \\
    1 & SCM & Sign & \textbf{0.0379} & 0.144$^\dagger$ & 9 & 60 & 90ns\\
    \hline
    
    \multicolumn{8}{p{190pt}}{*Theoretical, $\dagger$Different FPGAs used}, 
    \end{tabular}%
     
  \label{tab:SCNSCM}%
\end{table}%
In this section, we compare the FPGA SCN model proposed in \cite{gao2020fpga} using a piece-wise linear approximation of the sigmoid function. The FPGA in this paper uses a clock speed of 766MHz, so the values are evaluated at this speed and 100MHz. Table \ref{tab:SCNSCM} shows the difference in performance between the SCN model and the two models with the same number of hidden nodes; further, two models with 60 nodes are also shown. In this comparison, we can see that our FPGA SCM model performs two times quicker than the FPGA SCN model; it also outperforms using either sign or step activation with the same number of nodes. Further, we can see that at 60 nodes, it far exceeds the performance of the FPGA SCN model, both in speed and accuracy. This demonstrates that the SCM algorithm is more suitable for hardware implementation than SCN.

\section{Conclusion}
In this paper, we have demonstrated that SCM can be applied to hardware implementation for both single-layer and deep models. Further implementation using fixed point values, encoding schemes, activation functions and binary inputs is provided. The results demonstrate that the encoding schemes can provide a suitable representation of the models input. The performance results show that the hardware can provide outputs very similar to what is calculated on a PC and closely match the target. In terms of memory, it is shown that a large memory reduction is achieved using the hardware solution. And in terms of time, single-layer models can be evaluated in around 100ns and deep models in around 240ns. Compared to the existing hardware solution using SCN, our SCM FPGA implementation is faster and provides better accuracy. This work can be extended by exploring alternative encoding schemes, finding optimal parameters and using different mechanism models. Further, this work can be applied to real-world applications in industry. 

\section*{Acknowledgment}
We are grateful to Professor Xu Li from Northeastern University, China, for sharing two industrial datasets.

\bibliographystyle{IEEEtran}
\bibliography{IEEEabrv,SCM_FPGA}

\end{document}